\documentclass{article} % DO NOT CHANGE THIS
\usepackage{arxiv}  % DO NOT CHANGE THIS
\usepackage{graphicx} % DO NOT CHANGE THIS
\usepackage{amsmath}

\usepackage{booktabs} 
\usepackage{amsfonts}       % blackboard math symbols
\usepackage{algorithm}
\usepackage[noend]{algpseudocode}
\usepackage{breakurl}
\usepackage[switch]{lineno} 

\usepackage{cite}
\usepackage{float}
\usepackage{flushend}
\usepackage{multicol}
\usepackage{balance}
\usepackage{lipsum}
\usepackage{stfloats}

\graphicspath{ {./images/} }

\title{Evolutionary Selective Imitation:\\ Interpretable Agents by Imitation Learning Without a Demonstrator}

\author{Roy Eliya\\
School of Informatics\\ University of Edinburgh\\
\texttt{s1893676@ed.ac.uk}
  \And
J.~Michael Herrmann\\
School of Informatics\\ University of Edinburgh\\
 \texttt{michael.herrmann@ed.ac.uk}}
\date{}

\begin{document}
\maketitle
% \linenumbers 

\begin{abstract}
We propose a new method for training an agent via an evolutionary strategy (ES), in which we iteratively improve a set of samples to imitate: Starting with a random set, in every iteration we replace a subset of the samples with samples from the best trajectories discovered so far. 
The evaluation procedure for this set is to train, via supervised learning, a randomly initialised neural network (NN) to imitate the set and then execute the acquired policy against the environment.
Our method is thus an ES based on a fitness function that expresses the effectiveness of imitating an evolving data subset. 
This is in contrast to other ES techniques that iterate over the weights of the policy directly. 
By observing the samples that the agent selects for learning, it is possible to interpret and evaluate the evolving strategy of the agent more explicitly than in NN learning.
In our experiments, we trained an agent to solve the OpenAI Gym environment {\sc Bipedalwalker-v3} by imitating an evolutionarily selected set of only 25 samples with a NN with only a few thousand parameters. 
We further test our method on the Procgen game {\sc Plunder} and show here as well that the proposed method is an interpretable, small, robust and effective alternative to other ES or policy gradient methods. 
\end{abstract}
\twocolumn 
\section{Introduction}
Reinforcement learning (RL) agents are often effective at exploring their environment and find strategies that achieve high reward.
These strategies can be counter-intuitive or unexpected.
A core challenge of autonomous agent development is that it is difficult to deeply understand the strategy the agent discovered during training, before deployment to the real-world.

This problem is important because the strategies the agent discovered can be undesirable, exploitative or even dangerous in some point of their flow.
For example, in the game {\sc CoastRunners}, OpenAI demonstrated~\cite{CoastRunners} a racing agent that learned to exploit the game and achieve higher reward by crashing into other boats and repeatedly catching on fire, instead of actually finishing the race as the researchers intended.
This game demonstrates the exploitative and counter-intuitive nature of strategies that can be discovered by reinforcement learning agents.
In real-world scenario, the equivalent could be an automatic ship that learns to arrive faster by sailing in a way that dangers the passengers on board.
The concern of unexpected and dangerous agent behavior is growing in recent years as AI systems, and in particular robotic systems, become more widely adopted throughout the world.

The reason it is often difficult to fully understand trained policies is that the weights of a neural network (NN) are hard to interpret, which means researchers cannot rely on observing NN weights directly to understand the strategy the agent discovered.
Instead, researchers often rely on observing their policy in simulations in attempt to understand their agents better.
Simulations, however, often do not capture all of the possible states an agent would arrive at in the real-world, and relying on simulations is often not good enough to ensure the policy would not perform potentially damaging actions in the real-world.

Previous work attempted to help manage this problem.
Incorporating human feedback in the reward loop~\cite{humanfeedback} helps to ensure the agent is rewarded by following reasonable strategies, however human feedback can be demanding on human labor and be a limiting on the exploration of the agent.
Visualizing NN weights, such as microscope by OpenAI~\cite{microscope}, can help researcher to understand their policies, however, especially in Reinforcement learning where strategies are often complex, this method is limited in its application.

In this paper, we show that using Evolutionary Selective Imitation (ESI) we can generate policies that can be more easily interpretable, relieving the reliance on the NN weights to understand the policies.
ESI is similar to other evolutionary strategies (ES), with the main difference of iterating over which samples are best to imitate, instead of iterating over the weights of the NN policy directly.
The chosen imitation samples set, analogous to an airport baggage scanner, allow us to peak into the mind of the agent, so to speak, and develop a better understanding of the strategy the agent will follow.
As an illustration, in the {\sc CoastRunners} example~\cite{CoastRunners} mentioned above, researchers could potentially have observed ahead of deployment that the imitated samples focus on the specific section of the track the ship had learned to exploit, instead of the entire track, or perhaps observed suspicious samples in the set such that involve a ship set on fire. 
In our experiments, we present our method on {\sc Plunder} and {\sc Biped} as an interpretable, robust and effective alternative to ES or Markov Decision Processes methods, such as policy gradient.
\begin{figure}[th]
\centering
\includegraphics[width=.96\columnwidth]{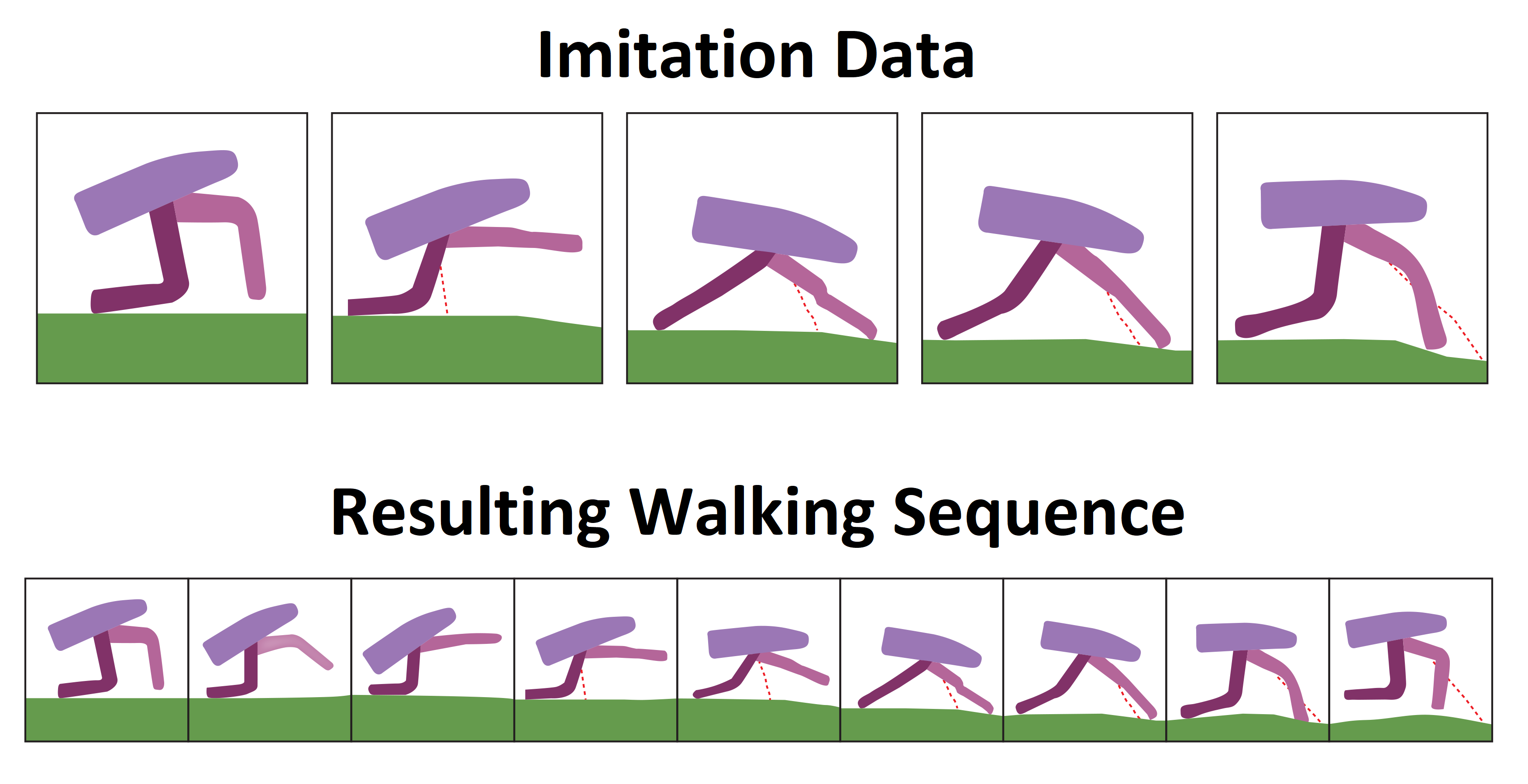}
\caption{Starting with random weights, our policy learned to walk in the {\sc Biped} problem after it was trained to imitate five samples. This set of five samples, each composed of a pair (observation, action), was iteratively mutated and improved in an evolutionary process of 40 million steps. The approximations of the five observations that have evolved, as taken from the walking sequence, are given under 'imitation data'. Below them is the walking sequence performed by the agent after it was trained on the five samples. We illustrate here that by observing the evolutionarily selected samples, we can develop better understanding of the strategy of the agent and the type of walk it will have when deployed.\label{figure1}}
\end{figure}

\section{Related Work}
\label{sec:headings}

\subsection{Interpretability}
Increasing the interpretability of agents in RL is an active research area and a variety of methods have been developed for that purpose~\cite{mythosOfInterpretability}. 
A comprehensive survey of visual interpretability of deep learning was carried out by~\cite{deepLearnInterpretabilitySurvey}. Ref.~\cite{programmaticallyInterpretable} introduced a method to produce, using a NN, programs that explain learnt policies, such that the policies are more easily interpreted, amendable and verifiable than neural networks.
Ref.~\cite{betterDeepQInterpretability} developed more easily interpretable architecture of deep Q-networks that achieved state-of-the-art training reward, however the features extracted were found to be shallow.
OpenAI offers a utility to visualize layers and neurons of a NN~\cite{microscope}, which can be used to analyze the features extracted from a NN.
Relying on interpreting NN directly, however, was demonstrated to be fragile to systematic perturbations~\cite{interpreteationFragile}, such that two visually indistinguishable inputs assigned to the same label can have very different interpretations by common interpretation methods.

% \twocolumn
\begin{figure*}[thb]
\centering
\includegraphics[width=.99\linewidth]{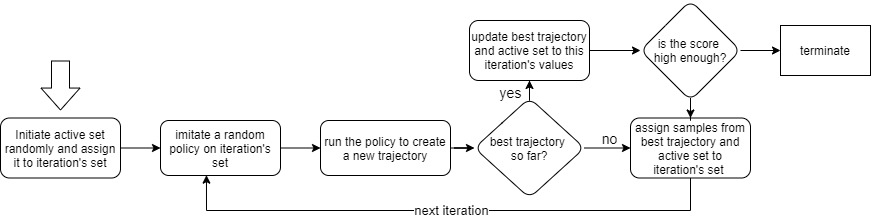}
\caption{Flow diagram of ESI. Starting with a random set of samples, new policies are generated by imitating a recursively improving set of samples, called \emph{active set}, which is used in training an agent towards a desired level of performance.\label{SchemeFig}}
\end{figure*}

Similarly to these methods, the interpretability of the control policy of an agent is an intended consequence of ESI.
However, ESI helps researchers understand their policy by developing through a selection of states that can be examined, relieving the reliance on the NN weights to understand the policy.

The gained interpretability of the policy allows us to better reduce the risk of the policy acting in damaging and unintentional manner after deployment. Specifically, Amodei et al.~\cite{concreteproblems} surveyed applications that lead to unintended and harmful behavior, and categorised these behaviors according to their origination. 
Using their categorization, our approach is intended to help to \emph{avoid negative side-effects} and \emph{reward hacking}, and to achieve \emph{scalable oversight} and \emph{robustness to distributional shift}. 

\subsection{Evolutionary Strategies}
Evolutionary strategies (ES) have been demonstrated as simple and effective methods in a variety of challenging RL tasks such as Atari games~\cite{atariNeuroevolution,geneticAlgorithmsCompetitive} and control of simulated robots~\cite{Ha2018designrl}.
Recently, ES~\cite{selfinterpretable} was used to evolve small agents that learned to solve vision tasks by directing attention to selectively chosen pixels. 
This selective attention also provide the benefit of interpretability, which can be gained by observing which pixels of the image the agent chose to attend to.
Similarly to these method, ESI uses evolutionary techniques to solve RL environments.
In contrast to these examples, however, ESI does not use the evolutionary strategy to improve the weights of the policy directly, but instead to improve a set of samples to imitate with imitation learning.

ES has similarities to ESI that we want to exploit as well as to extend here. 
One important benefit of ES was demonstrated by OpenAI ~\cite{evolutionarystrategies} when solving problems by distributing the computational load on a large cluster. 
Following this demonstration by OpenAI, ESI can similarly be distributed on a large cluster of machines.
Because each episode is composed of iterations that can run independently, by running all of the iterations of an episode in parallel, we can share the workload load between the machines and combine the results only once by the end of the episode.

\subsection{Imitation Learning}
Our method uses a form of imitation learning to develop the policy of the agent. 
In imitation learning, a.k.a.~learning from demonstration, a policy learns to generalize over a set of state and actions achieved through a demonstrator.
This method was found to be effective, among others, with self-driving vehicles~\cite{bojarskiSelfDriving,alvinnAuthCar} and robotic motion \cite{junBipedFromImitation,MrinalmotionFromImitation}.
A comprehensive review of the method can be found in~\cite{imitationlearningsurvey}.

There is a variety of methods that use imitation learning to improve the effectiveness of a learning agent.
Expert demonstrations were found to improve the efficiency of RL techniques and to solve tasks where RL alone is ineffective~\cite{NazirRLwithDemons}. 
These demonstrations were also shown to be used to improve the effectiveness of RL in the case where these demonstrations are imperfect~\cite{imperfectDemonstrations}.
Furthermore, combining imitation learning with interventions, such that the agent learns behaviour from a demonstration via imitation learning, and then leverage intervention data to improve further, was found to improve learning over imitation learning alone~\cite{humanDemoAndIntervention}.
A recent example of imitation learning to solve difficult tasks was demonstrated~\cite{singledemonstration} to learn a solution by studying a single expert demonstration. 
The main difference of our approach and other methods mentioned here, is that we do not require prior demonstration to start the imitation process.
Instead our policy starts with a trajectory generated by executing the random policy on the environment, and iteratively improves the imitation data through an evolutionary process.
Another difference to the method demonstrated by~\cite{singledemonstration} is that our method focuses on the beginning of the evolving solution that is then expanded towards the further stages of the behaviour, rather than learned backwards from the goal.

\section{Selective Imitation}
Selective imitation is motivated by learning of skill from demonstration in humans.
Even for complex demonstrations, and humans are able to choose important aspects and discernible aspects, and only these parts are then imitated, while distractors are ignored. 
For example, a young child can watch a professional soccer player on TV and try to imitate some of his moves with his playmates.
If the professional slips and loses the ball, the child will recognize the slip was a mistake and will not attempt to imitate that.
Even if the professional player does not perform a mistake, most of the time, he would perform actions that are too difficult and advanced for the child to learn, or that are not crucial for success, such as running with the ball or waiting for the ball to come. 
The child would know not to pay too much attention to this behavior. 

Inspired by such examples, we propose a method that uses an evolutionary algorithm to solve the \emph{credit assignment problem}, i.e.~how to learn which parts of a demonstration should be imitated. 
The advantage of ESI is that the set of imitated samples provides utility, which is more useful than a data-independent exploration strategy.
ESI begins randomly and improves upon a selection of samples chosen from the best trajectories encountered so far in the environment.
In every learning step ESI chooses the set that yields the policy that provides the best trajectory, and continues this process recursively to discover better samples to imitate, as will be further explored in the Method section.

\subsection{Interpretability by Implicit Data Selection}

Policies developed by ESI are more interpretable because, in contrast to the ``black box'' nature of the weights of a NN, the samples the policy chose to imitate can be observed.
This is important because the strategy the agent developed during training, can often be hard to anticipate by researchers (a.k.a. the control problem), and observing the agent in simulations is often not sufficient to ensure intentional behaviour and not accidentally overlook aspects of the strategy that can be dangerous or exploitative.
In analogy to a code review, observing the samples the policy imitated allows us to look inside the policy, understand the policies strategy better, and look for any suspicious or unintended samples.
In addition to preventing unintentional behavior, better understanding of the policy can also assist researchers to debug and improve sub-optimal performance. 
Researchers can also analyze specific training samples that caused sub-optimal behavior; when an undesirable behavior is observed, unlike black-box policies, a researcher can directly search the imitation data, determine what part is responsible and better understand the source of the behavior.

\subsection{Efficiency and Flexibility}

The resulting NN policy can often be small (thousands of parameters).
Because the NN is trained in a supervised fashion on a small number of samples, only a small NN is required to fully imitate the data.

This method can be used to improve upon an existing human demonstration. 
This can be done by initializing the method with an expert demonstration of solving the environment made by a skilled individual, instead of a randomly generated one.
In this way the algorithm would improve upon the existing demonstration to reach potentially higher results than could have been reached with a random initialization, while still remaining creative.

Each method episode can also be distributed among many independent machines, as demonstrated by OpenAI~\cite{evolutionarystrategies}.
In every episode, iterations run independently from one another, and as such, these iterations can be parallelised effectively in a computer cluster.

\section{Methods}
We use an iterative, evolutionary approach to find the best set of samples to imitate ${A} = ({s}_{j}, {a}_{j}, {r}_{j})_{j=0}^{M}$, called the active set. 
In every iteration the active set is mutated to produce a new set of samples ${Q}_{i}$, which is imitated using supervised learning to teach a neural network policy ${\pi}_{i}(\theta)$ to be executed on the environment and create a trajectory ${T}_{i}=({s}_{j}, {a}_{j}, {r}_{j})_{j=0}^{J}$.
The overall best trajectory encountered in any iteration thus far $\tilde{T}=(\tilde{s}_{j}, \tilde{a}_{j}, \tilde{r}_{j})_{j=0}^{J}$, as measured by the total reward sum $\sum_{j=0}^{J} \tilde{r}_{j}$, is recorded.

\begin{table}[ht]
\begin{tabular}{|c|l|}
\hline
$M$             & Size of active set                           \\ \hline
$N$             & Number of iterations per episode             \\ \hline
$\lambda$             & Percentage of samples replaced every mutation             \\ \hline
$P$             & Number of samples replaced every mutation \\ \hline

$A$             & Active set                                   \\ \hline
$A'$             & Subset of the active set                                   \\ \hline
$\tilde{T}$      & Best trajectory thus far \\ \hline
$\tilde{T}'$      & Subset of the best trajectory thus far \\ \hline
$e$             & Episode number                               \\ \hline
${L}_{e}$  & Sampling limit of episode                      \\ \hline
$i$             & Iteration number                             \\ \hline
${Q}_{i}$  & Imitation data of iteration                   \\ \hline
${\pi}_{i}(\theta)$ & Iteration policy                           \\ \hline
${T}_{i}$  & Iteration policy execution trajectory                        \\ \hline
$j$             & Step number                                  \\ \hline
${s}_{j}$             & State                                         \\ \hline
${a}_{j}$             & Action                                       \\ \hline
${r}_{j}$             & Reward                                        \\ \hline
\end{tabular}
\caption{List of variables and parameters.}
\end{table}

The size of the active set is $M$ and it is mutated such that some samples are replaced with new samples from the best trajectory, as explained in Fig.~\ref{SchemeFig}.
The mutation of the active set in every iteration ${Q}_{i}$ is the result of union ${Q}_{i} = A'\cup\tilde{T}'$ of samples from the active set $A'$ and samples from the best trajectory $\tilde{T}'$.
$A'$ is composed of $M-P$ randomly selected samples from the active set $({s}_{t}, {a}_{t}, {r}_{t})_{t=0}^{M-P} \subseteq A$. $\tilde{T}'$ is composed of $P$ randomly selected samples from the best trajectory $({s}_{l}, {a}_{l}, {r}_{l})_{l=0}^{P} \subseteq \tilde{T}$, where $P = M \cdot \lambda$.

In every iteration, a policy uses imitation learning to imitate the active set.
That is, the training of the policy $\pi_{\theta}$ is done with supervised learning, such that the policy learns to minimize the error between the predicted action of a state $\pi_{\theta}\left({s}_{t}\right)$, to the demonstrated action of the state ${a}_{t}$.
The formula for the error would thus be $\sum_{t=0}^{T}\left\|\pi_{\theta}\left({s}_{t}\right)-{a}_{t}\right\|^{2}$.
The weights of the policy would be initialised randomly in every iteration and trained to optimize this error. 
In our experiments, we used a 1-hidden layer CNN for {\sc Plunder} environment, and 1-hidden layer simple NN for {\sc Biped} as the agent.

In order to improve over the course of the entire task, the agent must combine information learned in previous steps with new information which means here to explore new samples for imitation learning. 
To balance exploration and iterative improvement, in every iteration the algorithm mutates the active set by keeping only some of the samples and replacing the other with new samples from the best trajectory found until now. 
The best trajectory from which new samples are taken is updated regularly throughout the execution (see Fig.~\ref{SchemeFig}) and also provides new samples to be imitated.
This balance encourages the disregard of some of the less useful, or perhaps damaging, samples, which are replace by new samples, while size of active set is kept constant.

The active set is only updated once every episode, that consists of iterations in which the active set is mutated and evaluated.
By updating the active set only once, it can be explored thoroughly during an episode before it is updated based on the best mutation found.
This prevents greedy behavior that, in our experiments, often leads the agent to a local optimum.
This happened when slightly better sets of samples are greedily chosen after every iteration, until no further improvement was possible by replacing the subset. 
In other words, too many samples were chosen that produce slightly better but still sub-optimal behavior, and the dominance of these samples would prevent a qualitatively better strategy to emerge. 
Instead of updating the active set after every iteration, one active set is explored in depth for many iterations in an entire episode.
Then, the best set found is chosen as as the active set for the next episode.
The number of iterations per episode is set by the hyperparameter $N$. Obviously, a higher number of iterations reduces the risk of the model to be stuck in a local optimum, at the cost of higher computational requirements.

In order to improve learning, samples in every iteration are not taken from the entire length of the best trajectory, but only from the initial ${L}_{e}$ observations, i.e.~from $(\tilde{s}_{j}, \tilde{a}_{j}, \tilde{r}_{j})_{j=0}^{{L}_{e}}$. 
For example, in the {\sc Biped} experiment, we sample initially only from the first five samples of the trajectory.
Because this initial scope is smaller than the samples in the imitation data ${L}_{0} < M$, there would be repetition of samples in the imitation data in the first episode. 
In every episode $e$ the length of the scope ${L}_{e}$ from which samples are taken is increased by 1. 
Limiting the sampling scope to the initial part of the trajectory and increasing the scope gradually improves the effectiveness of the sampling by focusing to the relevant direction in the data space, such that it becomes more likely to identify sequential relations in a complex search space that can be assumed here to have a causal structure. 
Limiting the samples to the beginning of a trajectory turned out to be critical for the success of the our method in both of the experiments in {\sc Plunder} and {\sc Biped}. 

The complexity of the policy is controlled by a hyper\-parameter $M$ (see Table 1) that sets the number of samples in the active set that is imitated by the policy.
A larger number of samples can in principle support the formation of more complex strategies such that the agent can succeed in more difficult environments, but this increases computational cost in training, an increased risk of overfitting, and more human labor required to fully examine the imitation data.

\begin{algorithm}[H]
\caption{Evolutionary Selective Imitation}
\begin{algorithmic}[1]
\Procedure{Main}{}
\Ensure Hyperparameters set
\State Initiate $M$, $N$, ${L}_{0}$, $\lambda$ 
\Ensure Variables initiated
\State $bestTraj$ $\gets$ random trajectory  $\tilde{T}$ 
\Comment{best so far}
\State ${L}_{e}$ $\gets$ ${L}_{0}$ 
\State $activeSet$ $\gets$ random subset of $bestTraj$ of size $M$
\While {Creating new episodes} 
\State ${L}_{e}$ $\gets$ ${L}_{e-1} + 1$ 
\Comment{\text{ Increase scope }} 
\While {running $itersPerEpisode$ iterations}
\State {\sc MutateSet} $\gets$ $M, N, L_e, \lambda, \tilde{T}, A$
\State $imitationData$ $\gets$ $MutateSet$
\State Train ${\pi}_{i}(\theta)$ on $imitationData$
\State \textbf{Execute} $\pi_{i}(\theta)$  \Comment{run iteration policy}
\State $iterationTraj$ $\gets$ $({s}_{j}, {a}_{j}, {r}_{j})_{j=0}^{J}$ 
\If { 
$\sum_{j=0}^{J} r_{j} > %
\sum_{j=0}^{J} \tilde{r}_{j} $} 
\State $bestTraj \gets iterationTraj$
\State $bestPolicy \gets {\pi}_{i}(\theta)$
\State $bestPolicyData \gets imitationData$
\EndIf
\EndWhile
\Comment{\text{Update and focus}} 
\State $activeSet \gets bestPolicyData$ 
\EndWhile
\State \Return {$bestPolicy$} 
\EndProcedure
\end{algorithmic}
\end{algorithm}

\begin{algorithm}[H]
\caption{MutateSet}
\begin{algorithmic}[1]
\State \textbf{Summary}: Replaces part of $activeSet$ with samples from $bestTraj$ and returns new set.
\Procedure{MutateSet}{$M, N, L_e, \lambda, \tilde{T}, A$} 
\State $P$ $\gets$ $M \cdot \lambda $ \Comment{mutation size}
\State $new \gets$ random subset of $(\tilde{s}_{j}, \tilde{a}_{j}, \tilde{r}_{j})_{j=0}^{{L}_{e}}$ of size $P$
\State $kept \gets$ random subset of $A$ of size $(M-P)$
\State $newActiveSet \gets$ $new$ $\cup$ $kept$ 
\State \Return $newActiveSet$
\EndProcedure
\end{algorithmic}
\end{algorithm}

\section{Experiments}

The goal of our experiments is the study of the effectiveness and the robustness of policies that are generated by ESI.
We also want to find out whether the behavior of the agent is predictable based on the data that it has chosen to imitate.
Lastly, we are going to test the importance of factors, such as exploration strength and active set size, on the results. 
We evaluated the method on two tasks, namely: {\sc Bipedalwalker-v3} ({\sc Biped} for short) and {\sc Plunder} (Easy difficulty) from Procgen (environment from OpenAI). 
Hyperparameters will have to be chosen differently in order to account for the specificity of each game.

\subsection{Hyperparameters}

    \begin{table}[ht]
    \centering
    \begin{tabular}{c|c|c}
                        & {\sc Biped} & {\sc Plunder} \\ \midrule
     $M$         & 25    & 160     \\
     $N$ &125   & 30      \\
    ${L}_{0}$ & 5   & 20      \\
    
    $\lambda$ & 0.5   & 0.5      \\
    \midrule
    $K$& 15    & 15     \\
    $\eta$& 0.005     & 0.005       \\
    $T$& 200   & 200      \\
    \end{tabular}
    \caption{Hyperparameters used in our experiments  ({\sc Biped} and {\sc Plunder}). $K$ (batch size), $\eta$ (learning rate) and $T$ (number of back propagation iterations) refer to the hyperparameters used in the imitation neural network, which were similar for {\sc Biped} and {\sc Plunder}.\label{Hypertable}}
    \end{table}
    
    To create agents that imitate the samples in every iteration (25 in {\sc Biped} and 160 in {\sc Plunder}) (see Table 2), we used a simple single-hidden-layer neural network. 
    In {\sc Biped} we used 40 hidden nodes for a total of 2,804 parameters. 
    In {\sc Plunder} we used a CNN with 128 nodes in the hidden layer for a total of 58,539 parameters. 
    Achieving good scores with such small number of parameters demonstrates an advantage of learning strategies with imitation learning on a few selected samples. 
    
    The range of hyperparameters tried is described in Figs.~\ref{Reward_biped}\nobreakdash-\ref{MutationStrength8} and the final hyperparameter setting was chosen to maximize reward in the respective game, under computational constraints. To run the experiments we use a 64-bit Windows 10 computer with Intel(R) Core(TM) i7-10510U CPU @ 1.80GHz, 16.0 GB RAM, and  the software Python 3, Pytorch v1.4.0, NumPy v1.18.1. 

\begin{figure}[ht]
\centering
\includegraphics[width=.66\columnwidth]{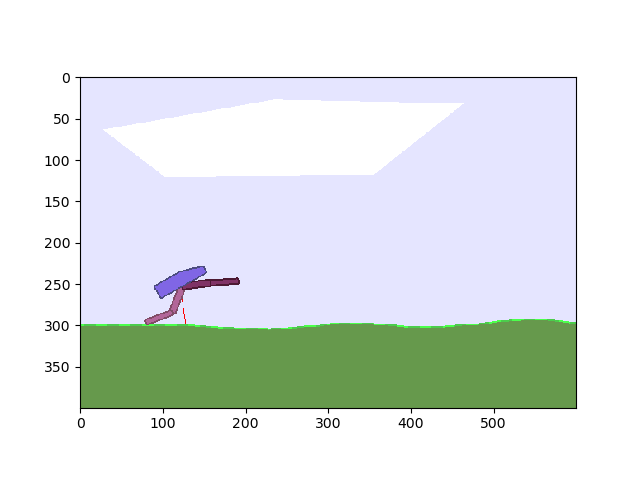}
\caption{Screenshot of Biped environment. The robot is tasked to reach the end of the route. Red line is a visualization of the Lidar sensor.\label{BipedScreenshot}}
\end{figure}

\subsection{Bipedal Walker}

\begin{figure}[ht]
\centering
\includegraphics[width=.99\columnwidth]{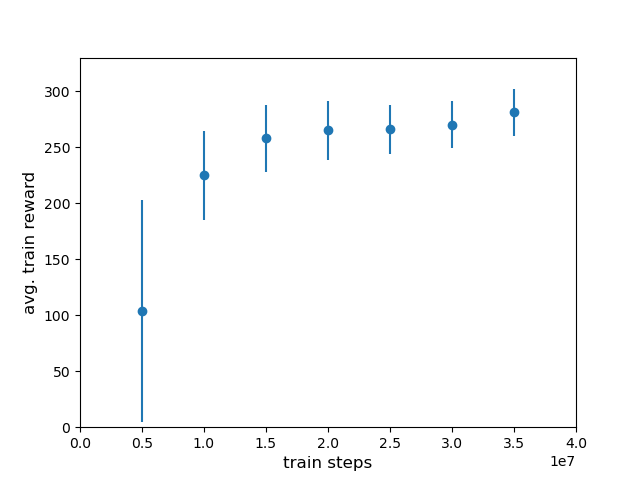}
\caption{Average and standard deviation of 10 experiments expressing the reward as function of training steps (millions). Parameters as given in Table~\ref{Hypertable} \label{bipedTrainingSteps}.}
\end{figure}

In {\sc Biped} (OpenAI Gym v0.17.0), the aim is the control of a robot, as shown in Fig.~\ref{BipedScreenshot}, and to reach the end of a route by controlling the velocity of the knee and hip joints (4 DoF) of a simple robotic walker. 
The reward is calculated based on the distance traverse by the robot. Reaching the end incurs +300 points, and falling leads to -100 points.
Applying torque costs small number of points, such that the choice of the gait type is important for achieving a higher score.
Solving {\sc Biped} is defined as achieving an average reward of 300 points over 100 consecutive trials. 
Our experiments (see Fig.~\ref{bipedTrainingSteps}) of running the algorithm for 40 million steps developed a policy that achieved an average reward of $300.02 \pm36.99$ over the required 100 trials.
We chose this environment in order to demonstrate ESI in a robotic environment and show how robotic movement can be developed effectively by ESI and predicted as a function of the sample set chosen for imitation. 

\begin{figure}[ht]
\centering
\includegraphics[width=.99\columnwidth]{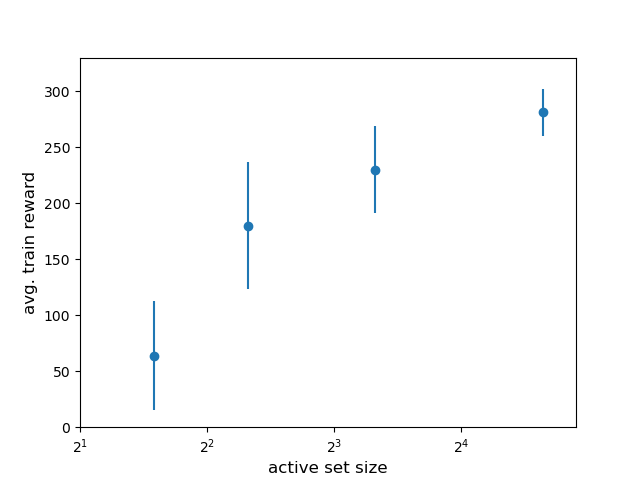}\hfill 
\caption{Average and standard deviation of 10 experiments expressing the reward after 40 million training steps on {\sc Biped} as function of the active set size, i.e.~the number of observations that are imitated by the policy.\label{Reward_biped}}
\end{figure}

As the agent is trained on more training steps, as demonstrated in Fig.~\ref{bipedTrainingSteps}, the reward increases as result of better data to imitate being discovered. The variance of the reward is large in the first few million of training steps as result of variance in the number of steps it requires to learn a simple walking sequence. 

In Fig.~\ref{Reward_biped} we show the effect of the active set size on the effectiveness of the agent. As we increase the active set size, more pairs $\left({s}_{j}, {a}_{j}\right)$ are imitated which entails the development of more complex strategies. 
This, however, exposes the agent to the risk of overfitting and increases the computational demands for an effective training of the policy.

We also demonstrate that the proposed method can develop an agent that achieves a score of $217.2 \pm4.99$ in average of 100 consecutive runs in {\sc Biped} by imitating only five samples.
Each sample is composed of a pair $\left({s}_{j}, {a}_{j}\right)$. 
In Fig.~\ref{figure1}, we present the graphical representation of all the states in the sample set.
With the knowledge that these are all the samples the agent chose to imitate; we know this set of samples encompasses the entire strategy.
Through this set, we can interpret the agent better before deployment.
This is thus an example of the benefit of interpretability that is inherent in using ESI.

\begin{figure}[htb]
\centering
\includegraphics[width=.48\columnwidth]{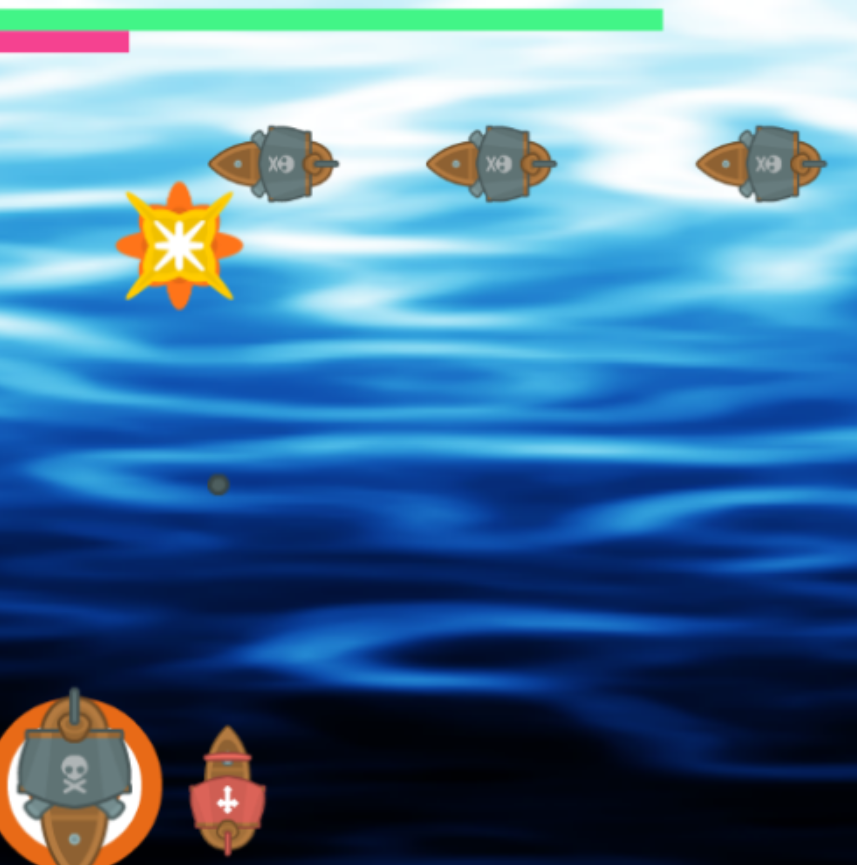}
\caption{Screenshot of Plunder environment. Circle image in lower left corner signifies the type of the target ships, and the ship above are the potential targets to destroy.\label{PluderScreenshot}}
\end{figure}

\begin{figure}[htb]
\centering
\includegraphics[width=.99\columnwidth]{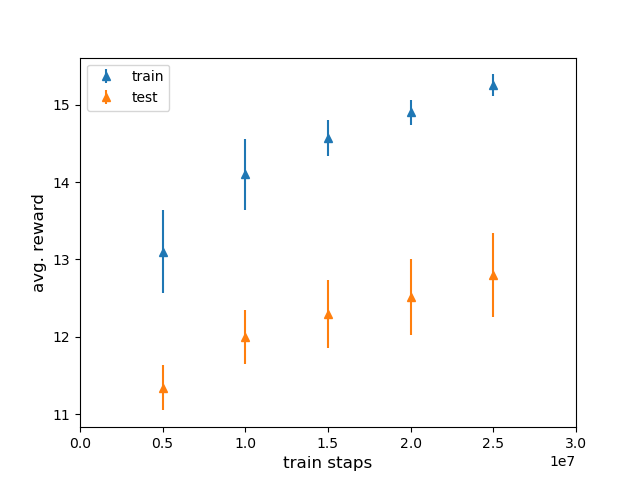} \caption{Average and standard deviation of 10 experiments expressing the reward in {\sc Plunder} as function of training steps in millions (Blue: training rewards, orange: testing reward). \label{RewardPlunder6}}
\end{figure}

\begin{figure}[ht]
\centering
\includegraphics[width=.99\columnwidth]{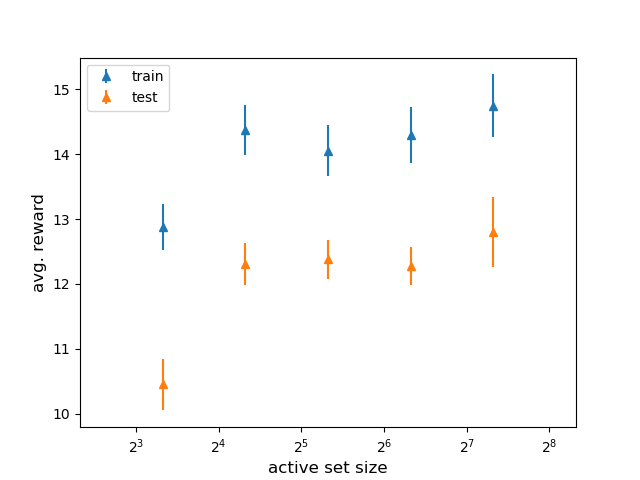}
\caption{Average and standard deviation of 10 experiments expressing the reward in {\sc Plunder} as function of active set size, i.e.~the number of observations the policy trains on (Blue: training reward, orange: testing reward). Larger set size allows for more complicated strategies to evolve but increases the chance or overfitting.\label{ResultPlunder7}}
\end{figure}

\begin{figure}[!ht]
\centering
\includegraphics[width=.99\columnwidth]{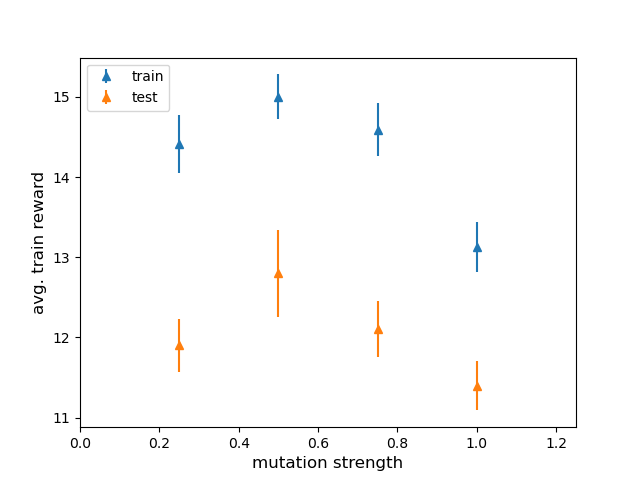}
\caption{Average and standard deviation of 10 experiments expressing the reward as function of the mutation strength, i.e.~the ratio of samples replaced per iteration. Higher percentage indicates more exploration.\label{MutationStrength8}}
\end{figure}

\subsection{Plunder}
We chose to evaluate the agent on {\sc Plunder} (Difficulty: Easy) environment of Procgen (OpenAI package Procgen v0.9.4 ~\cite{procgen}) for the purpose of assessing the generalization of the agent and to measure its effectiveness in a pixel-based environment.
Procgen is a benchmark developed by OpenAI that helps to measure the robustness of an agent by separating training and evaluation to different sets of procedurally generated levels. 
In environments that are part of Procgen, the agent first trains on a limited number of levels, 200 in this paper, for several iterations and then evaluated on a different set of 1,000 levels to gauge not only the effectiveness of the strategy it developed, but also the ability of the agent to not overfit to the 200 training levels.

In {\sc Plunder}, a player is tasked, under time constraints, to control a ship with the aim of destroying target ships specified by a symbol in the bottom-left corner, as shown in Fig.~\ref{PluderScreenshot}.
Levels are different from each other by the color and symbol of the target ship and the order and number of their appearance in the game.
The reward evaluation metric is calculated based on the number of target ships destroyed and the number of friendly ships that were not shot at.

Our agent was trained on 200 training levels and achieved an average score of 12.8 when evaluated on 1,000 unseen levels (see Fig.~\ref{RewardPlunder6}). 
In comparison, OpenAI used PPO to achieve 5.07 as the baseline result in Ref.~\cite{procgen}.
This result, in comparison to the baselines in Procgen paper, demonstrates the robustness that comes from using simple policies developed by ESI on the {\sc Plunder} environment, however, we found that we could not replicate these results on some other games in Procgen, such as Starpilot or Bossfight.

In addition, we tested {\sc Plunder} for how well the policy does as function of the active set size it trains on (see Fig.~\ref{ResultPlunder7}). 
There is a general increasing trend of reward as the policy becomes more complex by imitating a larger set of samples.
However, as mentioned in the methods section, more complex strategies are harder to understand, can overfit the training set and require more computations in training, so that it is recommended to keep the active set smaller.

We also tested to measure how well the policy does as a function of the level of exploration.
In every iteration, the policy decides to drop a random subset of the samples it imitates and replaces them with new samples. 
The larger the subset it drops, the higher the 'mutation strength' is, the higher the exploration of the policy is. 
This has resemblance to the noise factor in evolutionary strategies; high value would mean more exploration at the cost of higher difficulty to converge effectively and precisely.
The reward as function of the percentage of samples replaced in every iteration is given in Fig.~\ref{MutationStrength8}.
As we can see, a combination approach of exploration (replacing some of the samples) and exploitation (saving the best samples gathered thus far from previous episodes) achieves a higher score. 
The trade-off peaks at 0.5, which means replacing half the samples in every iteration. 

\section{Conclusion}
We have proposed a training method that enables inspection of the strategy that an agent discovers during training, by providing a set of critical samples from the agents behaviour. This inspection can help to gather insight and to interpret the learnt strategy.
Our method employs imitation learning to imitate a selected set of samples, but it does not require any demonstration to mimic.
Instead, the agent recursively iterates over a set of samples and, starting from a random trajectory, evolves a specific, small set of imitation data to train on. It is able to maintain a good learning process by the evolutionary choice of data, which appears to be an effective method of data selection for imitation.
As demonstrated, by imitating only 25 samples ESI solved the {\sc Biped} control task, and by observing this data we can develop an understanding of the strategy of the agent without the need to rely on simulated experiments.
We further demonstrated in the {\sc Plunder} game that the resulting agent is effective and robust, and requires only a few thousand parameters.

% \onecolumn
\bibliography{references}

\begin{thebibliography}{10}

\bibitem{concreteproblems}
Dario Amodei, Chris Olah, Jacob Steinhardt, Paul Christiano, John Schulman, and
  Dan Mané.
\newblock Concrete problems in ai safety.
\newblock {\em arXiv preprint arXiv:1606.06565}, 2016.

\bibitem{imitationlearningsurvey}
Brenna~D. Argall, Sonia Chernova, Manuela Veloso, and Brett Browning.
\newblock A survey of robot learning from demonstration.
\newblock In {\em Robotics and Autonomous Systems}, pages 469--483, 2009.

\bibitem{bojarskiSelfDriving}
Mariusz Bojarski, Davide~Del Testa, Daniel Dworakowski, Bernhard Firner, Beat
  Flepp, Prasoon Goyal, Lawrence~D. Jackel, Mathew Monfort, Urs Muller, Jiakai
  Zhang, Xin Zhang, Jake Zhao, and Karol Zieba.
\newblock End to end learning for self-driving cars.
\newblock {\em arXiv preprint arXiv:1604.07316}, 2016.

\bibitem{microscope}
Shan Carter, Michael Petrov, and Ludwig Schubert.
\newblock Openai microscope, 2020.

\bibitem{humanfeedback}
Paul Christiano, Jan Leike, Tom Brown, Miljan Martic, Shane Legg, and Dario
  Amodei.
\newblock Deep reinforcement learning from human preferences.
\newblock In {\em Advances in Neural Information Processing Systems}, pages
  4299--4307, 2017.

\bibitem{CoastRunners}
Jack Clark and Dario Amodei.
\newblock Faulty reward functions in the wild, 2016.

\bibitem{procgen}
Karl Cobbe, Christopher Hesse, Jacob Hilton, and John Schulman.
\newblock Leveraging procedural generation to benchmark reinforcement learning.
\newblock {\em arXiv preprint arXiv:1912.01588}, 2020.

\bibitem{interpreteationFragile}
Amirata Ghorbani, Abubakar Abid, and James Zou.
\newblock Interpretation of neural networks is fragile.
\newblock In {\em Proceedings of the AAAI Conference on Artificial
  Intelligence}, pages 3681--3688, 2019.

\bibitem{betterDeepQInterpretability}
Amirata Ghorbani, Abubakar Abid, and James Zou.
\newblock Towards better interpretability in deep {Q}-networks.
\newblock In {\em Association for the Advancement of Artificial Intelligence},
  pages 4561--4569, 2019.

\bibitem{humanDemoAndIntervention}
Vinicius Goecks, Gregory Gremillion, Vernon Lawhern, John Valasek, and Nicholas
  Waytowich.
\newblock Efficiently combining human demonstrations and interventions for safe
  training of autonomous systems in real-time.
\newblock In {\em Proceedings of the AAAI Conference on Artificial
  Intelligence}, pages 2462--2470, 2019.

\bibitem{Ha2018designrl}
David Ha.
\newblock Reinforcement learning for improving agent design.
\newblock {\em Artificial Life}, 25(4):352--365, 2019.

\bibitem{atariNeuroevolution}
Matthew Hausknecht, Joel Lehman, Risto Miikkulainen, and Peter Stone.
\newblock A neuroevolution approach to general {A}tari game playing.
\newblock {\em Transactions on Computational Intelligence and AI in Games},
  6(4):355--366, 2014.

\bibitem{imperfectDemonstrations}
Mingxuan Jing, Xiaojian Ma, Wenbing Huang, Fuchun Sun, Chao Yang, Bin Fang, and
  Huaping Liu.
\newblock Reinforcement learning from imperfect demonstrations under soft
  expert guidance.
\newblock In {\em Association for the Advancement of Artificial Intelligence},
  pages 5109--5116, 2020.

\bibitem{MrinalmotionFromImitation}
Mrinal Kalakrishnan, Jonas Buchli, Peter Pastor, and Stefan Schaal.
\newblock Learning locomotion over rough terrain using terrain templates.
\newblock In {\em International Conference on Intelligent Robots and Systems},
  pages 167--172, 2009.

\bibitem{mythosOfInterpretability}
Zachary~C Lipton.
\newblock The mythos of model interpretability.
\newblock {\em Communications of the ACM}, 61(10):36--43, 2018.

\bibitem{NazirRLwithDemons}
Ashvin Nair, Bob McGrew, Marcin Andrychowicz, Wojciech Zaremba, and Pieter
  Abbeel.
\newblock Overcoming exploration in reinforcement learning with demonstrations.
\newblock In {\em 2018 IEEE International Conference on Robotics and Automation
  (ICRA)}, pages 6292--6299, 2018.

\bibitem{junBipedFromImitation}
Jun Nakanishi, Jun Morimoto, Gen Endo, Gordon Cheng, Stefan Schaal, and Mitsuo
  Kawato.
\newblock Learning from demonstration and adaptation of biped locomotion.
\newblock In {\em Robotics and Autonomous Systems 47}, pages 79--91, 2004.

\bibitem{alvinnAuthCar}
Dean~A Pomerleau.
\newblock Alvinn: An autonomous land vehicle in a neural network.
\newblock In {\em Advances in Neural Information Processing Systems}, pages
  305--313, 1989.

\bibitem{singledemonstration}
Tim Salimans and Richard Chen.
\newblock Learning montezuma's revenge from a single demonstration.
\newblock {\em arXiv preprint arXiv:1812.03381}, 2018.

\bibitem{evolutionarystrategies}
Tim Salimans, Jonathan Ho, Chen Xi, Szymon Sidor, and Ilya Sutskever.
\newblock Evolution strategies as a scalable alternative to reinforcement
  learning.
\newblock {\em arXiv preprint arXiv:1703.03864}, 2017.

\bibitem{geneticAlgorithmsCompetitive}
Felipe~Petroski Such, Vashisht Madhavan, Edoardo Conti, Joel Lehman, Kenneth~O.
  Stanley, and Jeff Clune.
\newblock Deep neuroevolution: Genetic algorithms are a competitive alternative
  for training deep neural networks for reinforcement learning.
\newblock {\em arXiv preprint arXiv:1712.06567}, 2017.

\bibitem{selfinterpretable}
Yujin Tang, Duong Nguyen, and David Ha.
\newblock Neuroevolution of self-interpretable agents.
\newblock In {\em Proceedings of the 2020 Genetic and Evolutionary Computation
  Conference (GECCO'20)}, pages 414--420. Association for Computing Machinery,
  New York, NY, USA, 2020.

\bibitem{programmaticallyInterpretable}
Abhinav Verma, Vijayaraghavan Murali, Rishabh Singh, Pushmeet Kohli, and Swarat
  Chaudhuri.
\newblock Programmatically interpretable reinforcement learning.
\newblock {\em arXiv preprint arXiv:1804.02477}, 2018.

\bibitem{deepLearnInterpretabilitySurvey}
Quan-shi Zhang and Song-Chun Zhu.
\newblock Visual interpretability for deep learning: {A} survey.
\newblock In {\em Frontiers of Information Technology \& Electronic
  Engineering}, pages 27--39, 2018.

\end{thebibliography}
\bibliographystyle{plain}

\end{document}